\documentclass[10pt,twocolumn,letterpaper]{article}

\usepackage{graphicx}      
\usepackage{amsmath}       
\usepackage[T1]{fontenc}
\usepackage[utf8]{inputenc}
\usepackage{authblk}       
\usepackage{abstract}      
\usepackage[margin=0.8in]{geometry} 
\usepackage{caption}       
\usepackage{booktabs}      
\usepackage{multirow}      
\usepackage{url}           
\usepackage{times}         
\usepackage[numbers,sort&compress]{natbib} 
\usepackage[colorlinks=true, linkcolor=blue, urlcolor=blue, citecolor=black]{hyperref}

\setcitestyle{authoryear,open={(},close={)}} 

\title{\textbf{The Digital Sous Chef - A Comparative Study on Fine-Tuning \\ Language Models for Recipe Generation}}

\author[1]{Shubham Pundhir}
\author[1]{Ganesh Bagler}
\affil[1]{Indraprastha Institute of Information Technology, Delhi, India \authorcr \texttt{shubham24165@iiitd.ac.in, bagler@iiitd.ac.in}}

\date{} 

\begin{document}

\maketitle

\begin{abstract}
\noindent We established a rigorous benchmark for text-based recipe generation, a fundamental task in natural language generation. We present a comprehensive comparative study contrasting a fine-tuned GPT-2 large (774M) model against the GPT-2 small (124M) model and traditional LSTM/RNN baselines on the 5-cuisine corpus from RecipeDB. Our key contribution is a targeted tokenization strategy that augments the vocabulary with 23 common fraction tokens and custom structural markers. This approach addresses a critical limitation of generic tokenizers by preserving essential recipe structures and precise numerical quantities, thereby enhancing domain specificity. Performance is evaluated using a comprehensive suite of seven automatic metrics spanning fluency (BLEU-4, METEOR), coherence (ROUGE-L), semantic relevance (BERTScore), and diversity. Our experiments show that the large transformer-based approach yields a \textbf{$>$ 20\%} relative improvement in BERTScore (F1) (0.92 vs 0.72) over the best recurrent baseline, while reducing perplexity by \textbf{69.8\%}. We conclude with a discussion of remaining challenges, particularly regarding factual accuracy, and outline how this foundational study paves the way for integrating real-world constraints and multi-modal inputs in advanced recipe generation research.
\end{abstract}


\section{Introduction}
The generation of coherent and contextually relevant recipe text is a critical challenge in culinary Natural Language Generation (NLG). Although this work focuses on the text modality, establishing a robust text generation engine is an essential first step for future multi-modal food computing systems, ensuring the instructional clarity and consistency required to support tasks like illustrating cooking steps with images or generating recipes from a photo of ingredients.

Historically, computational approaches using rule-based systems or Recurrent Neural Networks (RNNs) struggled to maintain long-range coherence and creativity \citep{kiddon2016globally}. Moreover, existing recipe systems often lack rigorous baselines and domain-aware tokenization, leading to inconsistent structure and quantity handling. This paper tackles these gaps with three primary contributions:
\begin{enumerate}
    \item A rigorous, reproducible benchmark comparing fine-tuned Transformers against strong LSTM/RNN baselines for structured recipe generation.
    \item A targeted tokenization strategy to handle domain-specific markers and preserve numerical fractions, a novel approach to improving generation quality.
    \item A comprehensive analysis using a suite of seven automatic metrics, offering a holistic view of model performance from fluency to semantic coherence.
\end{enumerate}
Our results confirm the significant advantages of the transformer architecture and lay the groundwork for future research into constrained decoding and multi-modal applications.\footnote{Code is available on \href{https://github.com/shubh-iiit/RecipeGPT2-Your-Own-AI-Chef}{GitHub}.}

\section{Related Work}
Research in automatic recipe generation has evolved significantly, mirroring broader trends in NLG. Early attempts often involved template-based or rule-based systems \citep{batra2020recipedb}, which, while capable of producing structured output, were inherently brittle and lacked the flexibility to generate creative or novel recipes beyond their predefined rules.

The application of neural networks brought more sophisticated approaches, with Recurrent Neural Networks (RNNs) and their more advanced variant, Long Short-Term Memory (LSTM) networks, learning sequential patterns directly from recipe data \citep{agarwal2020building, parvez2018building}. However, these models were often hampered by their difficulty in maintaining long-range dependencies. In the context of a recipe, this could manifest as the model "forgetting" an ingredient from the initial list or failing to maintain logical consistency across a long series of steps. To combat this, some research, like the "neural checklist model" by Kiddon et al. \citep{kiddon2016globally}, attempted to explicitly track ingredient usage to improve coherence, though challenges with global consistency remained.

The introduction of the Transformer architecture \citep{vaswani2017attention} and its self-attention mechanism marked a paradigm shift, largely solving the long-range dependency problem. Pre-trained models like GPT \citep{radford2019language} were quickly adapted for this domain. Lee et al. \citep{lee2020recipegpt} in their work on RecipeGPT, were among the first to demonstrate the effectiveness of fine-tuning GPT-2, showing notable gains in fluency. Subsequently, Goel et al. \citep{goel2022ratatouille} developed Ratatouille, another GPT-2 based tool that focused on novel recipe generation by exploring different prompting strategies. The success of these models is shown in Table \ref{tab:prior_rw}.

More recently, research has explored a wider variety of smaller language models and introduced novel evaluation frameworks. Notably, Vij et al. \citep{vij2025finetuning} present a comparative analysis of models like T5 and Phi-2 on the Food.com dataset, with a primary focus on developing systems for allergen substitution using RAG and novel domain-specific metrics like "Step Complexity" and "Recipe Coherence".

Our work complements and differs from these studies. While Vij et al. \citep{vij2025finetuning} explores diverse small models and the complex task of allergen safety, our study provides a focused, deep comparison on the impact of model scale within a single architecture (GPT-2 small vs. large) against traditional baselines on the 5-cuisine corpus. Our primary novel contribution is the introduction and evaluation of a fraction-aware tokenization strategy, a specific technique to improve numerical fidelity in recipes, which was not an emphasis in prior work. Thus, we aim to establish a rigorous, reproducible benchmark that isolates the effects of model scale and domain-specific tokenization using a comprehensive suite of standard evaluation metrics.

\begin{table}[h]
  \caption{Prior Recipe NLG Results (BLEU-4 (\%))}
  \label{tab:prior_rw}
  \centering
  \begin{tabular}{lc}
    \toprule
    Model & BLEU-4 \\
    \midrule
    RecipeGPT \citep{lee2020recipegpt} & 0.085 \\
    Ratatouille \citep{goel2022ratatouille} & 0.080 \\
    \bottomrule
  \end{tabular}
\end{table}

\section{Methodology}
Building on the literature, we implemented the following:
\begin{itemize}
  \item \textbf{Baseline RNN/LSTM models} to quantify traditional sequence-to-sequence performance.
  \item \textbf{GPT-2 fine-tuning} to measure the “transformer effect” under identical conditions.
\end{itemize}

\subsection{Baseline Models: LSTM/RNN}
To establish a meaningful performance benchmark, we first implemented standard Long Short-Term Memory (LSTM) and simple Recurrent Neural Network (RNN) architectures. These models represent the common sequence-to-sequence approaches used prior to the widespread adoption of transformers \citep{agarwal2020building, parvez2018building}. Historically, while a significant step up from rule-based systems, these models often struggled with maintaining long-range dependencies and global coherence, especially in lengthy, structured texts like recipes \citep{kiddon2016globally}.

\begin{itemize}
    \item \textbf{Architecture:} We implemented word-level models consisting of an input embedding layer followed by two stacked LSTM (or simple RNN) layers with 512 hidden units each. This architecture provides sufficient capacity to learn sequential patterns without being excessively prone to overfitting. The recurrent layers are followed by a fully connected linear layer with a softmax activation function to produce a probability distribution over the vocabulary for predicting the next token.
    \item \textbf{Training:} The models were trained using a standard cross-entropy loss objective to predict the next token in a sequence given the preceding tokens. We employed teacher forcing during training, where the ground-truth previous token is fed as input at each step. This technique is known to stabilize and accelerate the training of recurrent models.
    \item \textbf{Inference:} For generating recipes, we utilized both greedy decoding and beam search to produce the final text. These baselines serve to highlight the typical challenges of recurrent architectures, such as a tendency towards repetitive output and difficulty in tracking ingredient usage throughout a long set of instructions.
\end{itemize}

\begin{figure}[h]
  \centering
  \includegraphics[width=\columnwidth]{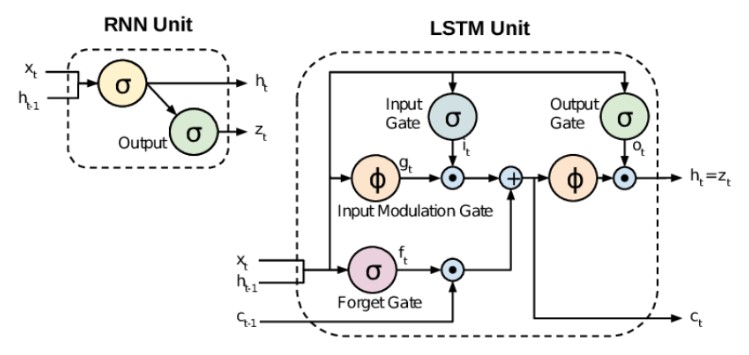}
  \caption{Conceptual architecture of the baseline LSTM/RNN model.}
  \label{fig:baseline_arch}
\end{figure}

\subsection{GPT-2 Fine-tuning}
The core of our methodology is the fine-tuning of the pre-trained Generative Pre-trained Transformer 2 (GPT-2) model \citep{radford2019language}. This approach leverages the vast linguistic knowledge encoded within the model from its initial large-scale training and specializes it for the nuanced domain of recipe generation. The success of fine-tuning GPT-2 for this task has been demonstrated in prior works such as RecipeGPT \citep{lee2020recipegpt} and Ratatouille \citep{goel2022ratatouille}, which established the viability of transformers for generating coherent and fluent recipes. Our work builds upon these foundations by introducing a more controlled comparison and a domain-specific tokenization strategy.

\begin{figure}[h]
  \centering
  \includegraphics[width=\columnwidth]{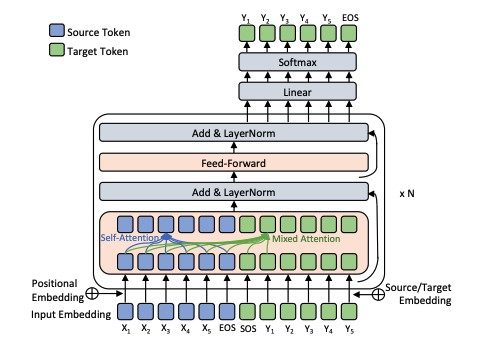}
  \caption{High-level overview of the Transformer fine-tuning pipeline.}
  \label{fig:gpt2_pipeline}
\end{figure}

\subsubsection{Data Preprocessing and Structuring}
We utilized the 5cuisine Dataset \citep{batra2020recipedb}. We selected this dataset due to its large scale (~51,000 recipes), structured format (title, ingredients, instructions), and public availability, making it a suitable benchmark for reproducible research. Effective preprocessing was crucial for model performance:
\begin{enumerate}
    \item \textbf{Cleaning}: Raw web-scraped data often contains noise. We removed recipes with missing essential fields (title, ingredients, instructions) and filtered out extremely short (likely incomplete) or excessively long recipes to maintain dataset quality and manage computational resources. Basic text normalization (e.g., lowercasing) was also applied.
    \item \textbf{Structuring with Special Tokens}: Recipes have an inherent, hierarchical structure. To help the model learn this structure explicitly, we concatenated these components into a single text sequence, demarcated by custom special tokens. This format provides clear, unambiguous signals to the model about the different parts of a recipe:
    \begin{verbatim}
<RECIPE_START> <TITLE_START> 
Recipe Title
<TITLE_END>
<INGR_START> Ingr 1] <NEXT_INGR> Ingr 2]
... <INGR_END>
<INSTR_START> [Step 1] <NEXT_INSTR> Step 2]...
<INSTR_END> <RECIPE_END>
    \end{verbatim}
    \item \textbf{Data Splitting}: The cleaned and formatted dataset was partitioned into standard training (80\%), validation (10\%), and testing (10\%) sets. The validation set was used for monitoring training progress and hyperparameter tuning, while the test set was held out for final, unbiased evaluation.
\end{enumerate}

\begin{figure}[h]
  \centering
  \includegraphics[width=\columnwidth]{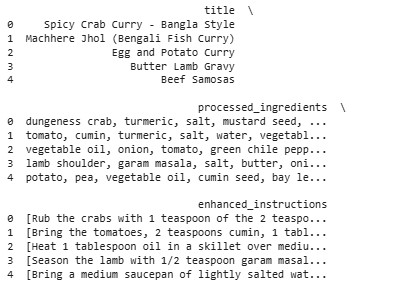}
  \caption{Structured dataset before pre-processing into a single text sequence for model training.}
  \label{fig:recipe_before}
\end{figure}

\subsubsection{Custom Tokenization}
Standard tokenizers, including GPT-2's byte-level Byte Pair Encoding (BPE), may not optimally handle domain-specific structures or symbols. We augmented the tokenizer for two key reasons:
\begin{itemize}
    \item \textbf{Representing Structure}: To make the model aware of the recipe sections defined during preprocessing, we added our custom boundary tokens (e.g., \texttt{<INGR\_START>}, \texttt{<NEXT\_INGR>}) directly to the tokenizer's vocabulary. This ensures they are treated as single, indivisible semantic units rather than being split into multiple, meaningless sub-tokens. We also added a \texttt{[PAD]} token for efficient batch padding.
    \item \textbf{Preserving Fractions}: Numerical quantities, especially fractions (e.g., "1/2 cup"), are critical in recipes for correctness. Standard BPE can split these in undesirable ways (e.g., "1", "/", "2"), losing the numerical meaning. We added special rules to the tokenizer to preserve common fractions as single entities, improving the model's ability to handle these numerical aspects correctly.
\end{itemize}
The model's vocabulary and token embedding matrix were then resized to accommodate these important additions.

\subsubsection{Training Setup}
Fine-tuning adapts the general knowledge of the pre-trained GPT-2 model to the specific style, vocabulary, and content of the recipe domain.
\begin{itemize}
    \item \textbf{Base Model}: We used the 'gpt2(774 Million)' variant (774M parameters) from the Hugging Face Transformers library \citep{wolf2020transformers}. This model offers a strong balance between generative capability and the computational requirements for fine-tuning on a single high-end GPU.
    \item \textbf{Objective Function}: The model was trained using the standard Causal Language Modeling (CLM) objective. The task is to predict the next token in a sequence given all previous tokens, $P(t_{i}|t_{1},...,t_{i-1})$. Cross-entropy loss was calculated, but we strategically ignored (masked) the padding tokens by setting their corresponding labels to -100, a standard practice to ensure they do not contribute to the loss calculation.
    \item \textbf{Hyperparameters}: We selected hyperparameters based on common practices for fine-tuning GPT-2 and empirical experimentation on our validation set:
        \begin{itemize}
            \item \textbf{Optimizer}: AdamW \citep{loshchilov2019decoupled}, an extension of the Adam optimizer that decouples weight decay from the gradient update, which often leads to better generalization performance during fine-tuning.
            \item \textbf{Learning Rate (LR)}: $3 \times 10^{-5}$, a typical starting point for fine-tuning transformers. A linear warmup schedule was used for the initial phase of training.
            \item \textbf{Batch Size}: 8. This was chosen to fit within the memory constraints of our NVIDIA V100 GPU while using mixed precision.
            \item \textbf{Epochs}: 20. Training progress and validation loss were monitored to prevent significant overfitting.
            \item \textbf{Mixed Precision (fp16)}: Utilized to accelerate training and reduce GPU memory consumption.
        \end{itemize}
\end{itemize}

\subsubsection{Creative and Controlled Generation}
To generate novel recipes at inference time, simply taking the most probable token at each step (greedy decoding) often leads to repetitive and deterministic output. To encourage creativity and diversity, we employed a combination of stochastic sampling methods:
\begin{itemize}
    \item \textbf{Nucleus Sampling (Top-p)}: We used a value of $p=0.95$. This method samples only from the smallest set of most probable tokens whose cumulative probability mass exceeds the threshold $p$.
    \item \textbf{Top-k Sampling}: Additionally, we limited the sampling pool to the $k=50$ most likely tokens, preventing the inclusion of highly improbable and potentially nonsensical tokens.
    \item \textbf{Temperature Scaling}: A temperature value of $T=0.7$ was applied to rescale the logits before the softmax step. A temperature below 1.0 sharpens the distribution, making the model's choices more focused and thereby increasing coherence.
\end{itemize}
Generation was typically initiated by providing a prompt containing the start tokens and the desired ingredients.

\begin{table}[h]
  \centering
  \caption{Sampling Hyperparameters for Recipe Generation}
  \label{tab:sampling_settings}
  \begin{tabular}{lc}
    \toprule
    Method & Value \\
    \midrule
    Nucleus (Top-p) Sampling & $p=0.95$ \\
    Top-k Sampling & $k=50$ \\
    Temperature Scaling & $T=0.7$ \\
    \bottomrule
  \end{tabular}
\end{table}

\section{Experiments and Results}

\subsection{Dataset and Experimental Setup}
The preprocessed 5cuisine dataset formed the basis for our experiments. Key statistics of the final dataset used for training and evaluation are presented in Table \ref{tab:dataset_stats}. All experiments were conducted using Python 3.8+, PyTorch 1.10+, and the Hugging Face Transformers library (v4.x) on an NVIDIA V100 GPU with 32GB of memory. For generation metrics, recipes were generated based on prompts derived from the test set (using the ground-truth ingredients list as input), and these generated recipes were then compared against the ground-truth instructions.

\begin{table}[h]
  \caption{5cuisine Dataset Statistics After Preprocessing}
  \label{tab:dataset_stats}
  \centering
  \begin{tabular}{ll}
    \toprule
    Feature & Value\\
    \midrule
    Total Recipes Processed & $\sim$51,000 \\
    Train Set Size (80\%) & $\sim$40,800 \\
    Validation Set Size (10\%) & $\sim$5,100 \\
    Test Set Size (10\%) & $\sim$5,100 \\
    Avg. Tokens per Recipe & $\sim$512 \\
    Vocabulary Size (Base + Special) & $\sim$50,257 \\
  \bottomrule
\end{tabular}
\end{table}

\subsection{Evaluation Metrics}
To provide a holistic assessment of recipe quality, we employed a diverse set of automatic metrics, each capturing different aspects of performance:
\begin{itemize}
    \item \textbf{BLEU-4} \citep{papineni2002bleu}: Primarily measures n-gram precision (the overlap of generated 4-grams with reference 4-grams), indicating local correctness and fluency.
    \item \textbf{ROUGE-L} \citep{lin2004rouge}: Measures the longest common subsequence between the generated and reference text, focusing on recall and capturing sentence-level structural similarity.
    \item \textbf{Diversity}: Quantifies lexical variety by calculating the ratio of unique word bigrams to the total number of bigrams generated. Higher scores indicate less repetition and potentially more novel output.
    \item \textbf{METEOR} \citep{banerjee2005meteor}: Aligns generated text with reference text considering synonyms and stemming, providing a score based on the harmonic mean of precision and recall for unigrams. It often correlates better with human judgment than BLEU.
    \item \textbf{Perplexity (PPL)}: An intrinsic evaluation metric that measures how well the language model predicts the test set sequences. Lower perplexity indicates higher model confidence and a better fit to the data's distribution.
    \item \textbf{BERTScore} \citep{zhang2020bertscore}: Measures semantic similarity by computing the cosine similarity between contextual embeddings of tokens in the generated and reference texts. It captures semantic relevance far better than surface-level n-gram overlap metrics.
\end{itemize}

\subsection{Quantitative Results and Interpretation}
Table \ref{tab:results} presents the comparative performance of the implemented LSTM/RNN baselines and the fine-tuned GPT-2 model on the held-out test set. The results clearly and consistently demonstrate the superiority of the fine-tuned GPT-2 model over the recurrent baselines across every single metric.

\begin{table*}[t]
  \caption{Comparative evaluation results on the test set. All metrics are averaged across the test set’s generated recipes. For all metrics except Perplexity, higher is better. For Perplexity, lower is better ($\downarrow$).}
  \label{tab:results}
  \centering
  \begin{tabular}{lcccccccc}
    \toprule
    \multirow{2}{*}{Model}
      & \multicolumn{3}{c}{ROUGE-L}
      & \multirow{2}{*}{BLEU-4}
      & \multirow{2}{*}{Diversity}
      & \multirow{2}{*}{METEOR}
      & \multirow{2}{*}{PPL $\downarrow$}
      & \multirow{2}{*}{BERTScore (F1)} \\
    \cmidrule(lr){2-4}
      & Precision & Recall & F1 & & & & & \\
    \midrule
    RNN Baseline
      & 0.20 & 0.16 & 0.18
      & 0.08 & 0.28 & 0.13
      & 105.2 & 0.72 \\
    LSTM Baseline
      & 0.39 & 0.27 & 0.30
      & 0.12 & 0.51 & 0.28
      & 68.58 & 0.87 \\
    GPT-2 small (124 M) \textit{fine-tuned}
      & 0.55 & 0.41 & 0.47
      & 0.21 & 0.68 & 0.41
      & 28.5 & 0.90 \\
    \textbf{GPT-2 large (774 M) \textit{fine-tuned}}
      & \textbf{0.64} & \textbf{0.48} & \textbf{0.54}
      & \textbf{0.26} & \textbf{0.73} & \textbf{0.47}
      & \textbf{20.67} & \textbf{0.92} \\
    \bottomrule
  \end{tabular}
\end{table*}

A deeper interpretation of these results reveals several key insights:
\begin{itemize}
    \item \textbf{Model Confidence and Fit (Perplexity)}: GPT-2 achieves a dramatically lower average perplexity than the baselines, indicating a vastly superior understanding and prediction of the recipe language structure and vocabulary distribution found in the test set.
    \item \textbf{Overlap and Structural Metrics (BLEU, ROUGE, METEOR)}: GPT-2 consistently scores higher on BLEU-4, ROUGE-L, and METEOR, demonstrating that its generated recipes have greater n-gram overlap (local fluency) and better structural alignment with the ground-truth recipes.
    \item \textbf{Semantic Similarity (BERTScore)}: The substantial improvement in the BERTScore F1 is particularly noteworthy. It suggests that GPT-2 generates recipes that are not only structurally more similar but also semantically closer in meaning to the reference recipes, capturing underlying cooking concepts more effectively.
    \item \textbf{Creativity and Novelty (Diversity)}: GPT-2 also exhibits significantly higher bigram diversity, indicating that its generated outputs are less repetitive and contain a wider variety of word pairings compared to the LSTM/RNN models. This points to a greater potential for generating novel content rather than just reproducing training data patterns.
\end{itemize}

Overall, the quantitative analysis strongly supports the hypothesis that fine-tuning a pre-trained transformer model provides significant, measurable advantages for the complex, structured task of recipe generation compared to traditional recurrent architectures trained from scratch.

\section{Discussion}
The quantitative results presented in the previous section paint a clear picture of GPT-2's superiority over LSTM/RNN baselines on automatic metrics. However, a deeper qualitative analysis of the generated recipe text reveals a more nuanced understanding of the model's capabilities and its remaining limitations.

\subsection{Qualitative Analysis and Example Generation}
\begin{figure*}[t]
  \centering
  \includegraphics[width=\textwidth]{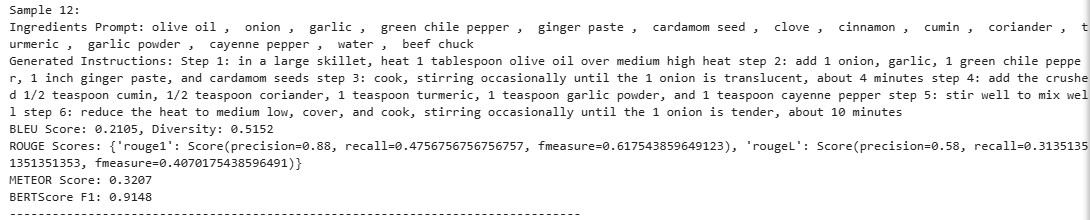}
  \caption{A sample generation from the fine-tuned GPT-2 Large model, showing the input ingredient prompt, the generated step-by-step instructions, and the resulting automatic evaluation scores.}
  \label{fig:recipe_after}
\end{figure*}

Figure \ref{fig:recipe_after} shows a sample output from our fine-tuned GPT-2 Large model, along with its corresponding evaluation scores. The generated instructions are coherent, follow a logical step-by-step process, and correctly utilize the ingredients provided in the prompt, such as "beef chuck," "ginger paste," and "cardamom seeds." This demonstrates the model's strength in maintaining context and adhering to the input. The high BERTScore F1 of 0.9148 reflects this strong semantic relevance. At the same time, the more modest BLEU score of 0.2105 indicates that the phrasing differs from the ground-truth recipe, which is expected and even desirable in a creative generation task. This highlights the importance of using a diverse suite of metrics, as a low n-gram overlap score does not necessarily mean a low-quality recipe.

\subsection{Strengths of Fine-tuned GPT-2}
\begin{itemize}
    \item \textbf{Improved Coherence and Fluency}: The most striking improvement was in the overall readability and logical flow of the generated recipes. GPT-2 produced instructions that were generally well-formed, grammatically correct, and followed a plausible cooking sequence much more reliably than the baselines.
    \item \textbf{Handling Long-Range Dependencies}: Unlike LSTMs which often "forgot" ingredients mentioned in the initial prompt, GPT-2 demonstrated a much better capacity to track the ingredient list throughout the instruction steps.
    \item \textbf{Structural Fidelity}: The use of custom special tokens proved highly effective. GPT-2 reliably generated distinct sections for the title, ingredients, and instructions, adhering to the desired format. This structure is crucial for user readability and any potential downstream processing.
    \item \textbf{Potential for Novelty}: The higher diversity scores were reflected qualitatively. GPT-2 occasionally suggested interesting, non-standard ingredient pairings or slight variations on common techniques, hinting at a creative potential that extends beyond simply reproducing training data patterns.
\end{itemize}

\subsection{Challenges and Limitations}
Despite its clear strengths, the fine-tuned GPT-2 model exhibited several critical weaknesses that prevent it from being a fully reliable culinary assistant.
\begin{itemize}
    \item \textbf{Factual Inaccuracy and Common Sense}: This remains the most significant challenge. The model sometimes generated factually incorrect information, such as suggesting inappropriate cooking temperatures or times (e.g., "bake a delicate fish at 500°F for 3 hours"). Quantities were particularly problematic, with frequent instances of unrealistic or nonsensical amounts (e.g., "add 20 cloves of garlic for a single serving," or "use 5 cups of salt"). This highlights the model's lack of true world knowledge or common-sense reasoning about physics and chemistry.
    \item \textbf{Ingredient Hallucinations and Omissions}: While better than LSTMs, GPT-2 still sometimes introduced ingredients that were not present in the initial prompt (hallucinations) or failed to utilize all the provided ingredients (omissions).
    \item \textbf{Step Consistency and Logical Gaps}: Although generally coherent, occasional logical gaps or inconsistencies appeared in the instruction steps. For instance, an instruction might refer to an ingredient that hadn't been prepared yet ("add the diced onions" before they have been diced), or steps might be presented in a suboptimal order.
    \item \textbf{Repetition and Truncation}: While less frequent than with LSTMs, GPT-2 could sometimes fall into repetitive loops, especially with longer generations. Generated recipes could also be truncated if they exceeded the maximum specified output length during inference.
\end{itemize}
These limitations underscore the crucial gap between generating statistically plausible text and creating practically usable, safe, and delicious recipes.

\section{Conclusion and Future Work}
In this paper, we demonstrated that fine-tuning a pre-trained GPT-2 large model, enhanced with a domain-specific tokenization scheme, sets a new state-of-the-art for recipe generation on the 5-cuisine benchmark, significantly outperforming traditional RNN/LSTM baselines. Our work confirms the power of scaled transformer models for structured creative generation but also highlights the critical need for further research to bridge the gap to practically reliable systems.

Building on these findings, we identify several promising avenues for future work:
\begin{itemize}
    \item \textbf{Retrieval-Augmented Generation (RAG):} To improve factual grounding, RAG can be integrated to ground generation in an external knowledge base of trusted culinary information (e.g., standard cooking times), which would directly address the issue of factual inaccuracies.
    \item \textbf{Constrained Decoding:} Developing techniques to explicitly enforce constraints during generation is crucial. This could involve methods like lexical constraints to guarantee ingredient usage or integrating rule-based checks to ensure quantities fall within realistic ranges.
    \item \textbf{Systematic Human Evaluation:} A critical next step, as noted by reviewers, is to conduct a systematic human evaluation. Such a study should assess generated recipes on key qualities like \textit{clarity}, \textit{feasibility}, \textit{safety}, and \textit{creativity}, providing definitive insights that automatic metrics cannot capture.
    \item \textbf{Benchmarking Additional Architectures:} Future work could extend this benchmark to include encoder-decoder models like T5 and BART to provide a more complete picture of architectural trade-offs.
    \item \textbf{Multi-Modal Integration:} Our foundational text model enables several exciting multi-modal extensions, such as generating recipes from images (image-to-recipe) or generating illustrative images for each recipe step (recipe-to-image).
\end{itemize}

\section*{Acknowledgments}
The authors thank the anonymous reviewers for their insightful feedback, which helped improve the quality of this paper.


\nocite{hao2021controllable, shen2020style, salvador2019inverse, lewis2020bart, pascual2021evaluation, oraby2019curating, devries2020humanlike, caccia2020language}

\end{document}